\documentclass{article}

\usepackage{arxiv}

\usepackage[utf8]{inputenc} 
\usepackage[T1]{fontenc}    
\usepackage{hyperref}       
\usepackage{url}            
\usepackage{booktabs}       
\usepackage{amsfonts}       
\usepackage{nicefrac}       
\usepackage{microtype}      
\usepackage{lipsum}
\usepackage{graphicx}
\usepackage{amsmath}
\usepackage{multirow}

\graphicspath{ {./images/} }

\title{Model Evaluation for Domain Identification of Unknown Classes in Open-World Recognition: A Proposal}

\author{
 Gusti Ahmad Fanshuri Alfarisy \\
  School of Digital Science | Department of Informatics \\
  Universiti Brunei Darussalam | Kalimantan Institute of Technology\\
  \texttt{20h8562@ubd.edu.bn} | \texttt{gusti.alfarisy@itk.ac.id} \\
   \And
 Owais Ahmed Malik \\
  School of Digital Science\\
  Universiti Brunei Darussalam\\
  \texttt{owais.malik@ubd.edu.bn} \\
  \And
 Ong Wee Hong \\
  School of Digital Science\\
  Universiti Brunei Darussalam\\
  \texttt{weehong.ong@ubd.edu.bn} \\
}

\begin{document}
\maketitle
\begin{abstract}
Open-World Recognition (OWR) is an emerging field that makes a machine learning model competent in rejecting the unknowns, managing them, and incrementally adding novel samples to the base knowledge. However, this broad objective is not practical for an agent that works on a specific task. Not all rejected samples will be used for learning continually in the future. Some novel images in the open environment may not belong to the domain of interest. Hence, identifying the unknown in the domain of interest is essential for a machine learning model to learn merely the important samples. In this study, we propose an evaluation protocol for estimating a model's capability in separating unknown in-domain (ID) and unknown out-of-domain (OOD). We evaluated using three approaches with an unknown domain and demonstrated the possibility of identifying the domain of interest using the pre-trained parameters through traditional transfer learning, Automated Machine Learning (AutoML), and Nearest Class Mean (NCM) classifier with First Integer Neighbor Clustering Hierarchy (FINCH). We experimented with five different domains: garbage, food, dogs, plants, and birds. The results show that all approaches can be used as an initial baseline yielding a good accuracy. In addition, a Balanced Accuracy (BACCU) score from a pre-trained model indicates a tendency to excel in one or more domains of interest. We observed that MobileNetV3 yielded the highest BACCU score for the garbage domain and surpassed complex models such as the transformer network. Meanwhile, our results also suggest that a strong representation in the pre-trained model is important for identifying unknown classes in the same domain. This study could open the bridge toward open-world recognition in domain-specific tasks where the relevancy of the unknown classes is vital.
\end{abstract}




\maketitle

\section{Introduction}\label{}

Most of the recognition models built upon deep neural networks have been evaluated with an assumption of closed-world settings. They assume that the evaluation of the empirical approximator (a model) using the test set sampled from the same distribution is sufficient. Many applications, e.g., medical waste classification \cite{Zhou2022ADL}, dog breed classification \cite{Wang2020DogDL}, and plant species recognition \cite{dyrmann_plant_2016} use this assumption and report high accuracy on the test set. Believing in the reported performance, especially the superior one in an open environment where unknown classes could present potentially a misleading portrayal of real-world performance. \cite{Boult_Cruz_Dhamija_Gunther_Henrydoss_Scheirer_2019} emphasized: "\textit{we recommend you do not trust a claim of intelligence that does not admit when it does not know or does not continue to learn}".

In a more realistic scenario, e.g. medical diagnosis \cite{Ge2021EvaluationOV} and human action \cite{Bao2021EvidentialDL}, a myriad of unknown image categories could be faced by a model. When a model is presented with unknown classes, one of the known classes will be assigned as the prediction output. The absence of rejecting unknown classes is a principal issue if the model will be utilized in the real world with a loosely controlled environment. The lab experimentation needs to shift the assumption in evaluating the model from a closed-world scenario to open-set recognition \cite{Scheirer2013TowardOS}. Moreover, the frequency of the unknown classes is likely abundant compared to the trained known classes. Providing a "skill" for any deep learning model to reject an unknown class makes the model a reliable machine.

\begin{figure*}[]
  \centering
  \includegraphics[scale=0.6]{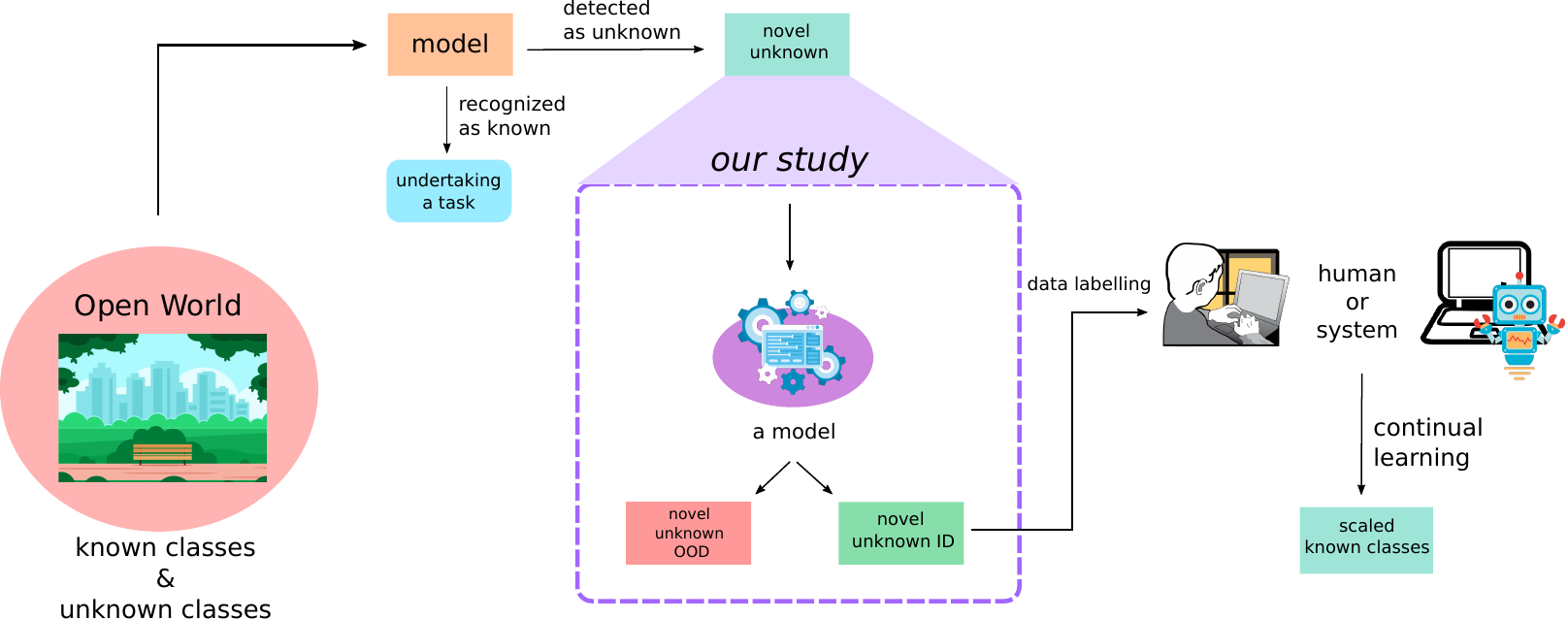}
  \caption{The position of our study as a bridge to open-world recognition in managing the unknown automatically.}
  \label{fig:ourstudy}
\end{figure*}

Open-world learning is an attempt to make a model "trustworthy". Open-world learning is a growing area that has an objective to elevate the model to have the ability to reject unknown classes and to learn continually from the rejected samples. Open-world learning is expected to collect the unknown instances and to discover the class by querying the human or by using an automatic system followed by the accumulation of the future knowledge gained by the model \cite{Chen2018LifelongML}. We can see open-world learning as a combination of three main components: open-set recognition, novelty discovery (in the case of automatic labeling), and continual learning. Open-set recognition is a solution for multi-class classification tasks to have a classification model with unknown rejection \cite{Geng2018RecentAI}. Meanwhile, continual learning is the study of expanding the knowledge in the deep learning models without catastrophically forgetting the previous knowledge \cite{lesort_continual_2020}.

In this research, we investigated the potential protocol for evaluating open-world learning in domain-specific tasks. As a model that manages specific tasks deployed in an open-world environment, a myriad of unknown classes can be experienced. Merely the related unknown classes that are relevant for a model that can be utilized for continual learning. Hence, separating the unknown into in-domain (ID) and out-of-domain (OOD) unknowns becomes imperative. Unfortunately, many open-set recognitions did not assess the performance of the dataset outside of the trained set, and open-world recognition did not have the capability of unknown separation.  

The position of the problem studied in this research can be seen in Figure \ref{fig:ourstudy}. In an open-world scenario, the model could reject the samples that do not belong to any known classes. The state of undertaking a normal task will be encountered when the samples are in known classes. The collected rejected data are managed by the separation of unknown ID and unknown OOD. The ID samples will be passed to the human or system to get the labels followed by incremental or continual learning to scale the model.

The task of separating the unknown is important to be executed. Since the model has an objective to solve a specific task on a particular domain, learning any rejected unknown that is not in the same domain could jeopardize the model. For example in the case of a waste robot that has a specific task to collect waste, a robot could encounter any arbitrary object leading to any type of unknown object including a non-waste object. Relying on the current assumption to accumulate the learning set based on rejected samples is not plausible as it contains unknown outside of waste categories. Hence, the identification of two different unknowns which are ID and OOD becomes imperative to be discussed.

Unfortunately, discriminating ID and OOD instances will confront imbalanced distribution as the OOD instances are in massive numbers or even infinite. It is hard to project any ID or OOD into a single probability density function. We need a piece of knowledge beyond what has been trained, outside of the known dataset.

A popular approach to utilize that knowledge in classification problems is through pre-trained models. They contain optimal parameters in solving large-scale classification tasks (e.g., ImageNet \cite{Deng2009ImageNetAL}). The parameters were updated to suit the performance on 1000 classes (in the case of ImageNet) which has been successfully employed to gain generalization for the target task in the closed-world scenario. However, it is unclear how these parameters could help to identify unknown ID and OOD instances. 

We utilized the pre-trained weights in pre-trained models in evaluating the problems of unknown separation. We employed the pre-trained models to learn two classes which are unknown ID and unknown OOD. We also investigate with the AutoML and Nearest Class Mean (NCM) classifier based on the clusters from latent in pre-trained models. We trained the models using the classes in ID and OOD and tested them with different classes of ID and OOD. Using this protocol, we can assess the robustness of any model that tackles a similar problem. Finally, we compared the approaches with the available deep novelty detection technique that utilizes an unknown set to detect the novelty as a fair comparison and results in poor performance when evaluating with our protocol.

Our contribution to the development of open-world recognition can be summarized as:

\begin{itemize}
    \item We proposed an evaluation for open-world recognition in domain-specific tasks for domain identification by using several domains.
    \item We experimented with three different approaches as a potential initial baseline to unknown separation for open-world recognition.
    \item We analyzed the effect of different pre-trained models using different approaches with the proposed protocols.
\end{itemize}

\section{Related Works}

To the best of our knowledge, there is no similar study to tackle the problem at hand, except our previous work that proposes a novel problem in OSR. \cite{Alfarisy2023} introduced and formalized the Open Domain-Specific Recognition (ODSR) that could reject unknown followed by the identification of unknown ID or OOD. Our work can be seen as one of the recognizers in ODSR that distinguish unknown ID from unknown OOD. In our approach, we incorporated the OOD set into the model and investigated the features in pre-trained models to achieve robust performance.

Meanwhile, the study by \cite{Wu2022UCOWODUO} addressed the separation of the unknown classes for open-world object detection problems, it was mainly focused on the assumption that these unknown classes exist in the same domain of interest, which is different from the problem presented here. Another study designed for detecting unknowns by \cite{Fontanel2022DetectingTU} did not emphasize the concern on the domain of interest.

Another field of study that is nearly similar is anomaly or novelty detection. The experimentation scenario also uses the same training distribution in the testing phase which separates known classes from unknown. In the study by \cite{Sabokrou2018AdversariallyLO}, the same distribution with training is drawn for testing in tackling novelty detection. The MNIST representing the digit domain was used for experimentation, their study took the samples from the first digit as the inliers and other digits as the outliers. Similarly, in a study conducted by \cite{Tack2020CSIND}, several benchmarking scenarios were investigated: unlabeled one-class dataset, unlabeled multi-class dataset, and labeled multi-class dataset. All of them use the same training distribution for the inlier. No assumption about the different characteristics of the inlier as the domain of interest was constructed that is possibly tested. \cite{Golan2018DeepAD} used the same settings in tackling anomaly detection problems by taking one label as a one-class dataset. \cite{Salehi2021AUS} compile the literature related to unknown detection which includes anomaly detection, novelty detection, open-set recognition, and out-of-distribution detection which also use similar experimentation to distinguish between known and unknown. Meanwhile, our study addresses a different problem that separating between unknown ID and unknown OOD classes.

\section{Evaluation Model and Methods}

\subsection{Problem Definition}

Our task is binary classification with a different objective. In conventional multi-class or binary classification tasks, the image's class in the same training distribution will be used for testing the performance. In contrast, our experimentation emphasizes testing the unknown label which is different from the training distribution but in the same domain of interest. To determine the belonging of the unknown label in the same domain, the model should have the capability of rejecting the unknown label that is outside of the domain of interest. In other words, we train the dataset and test it with other labels, e.g., in the food domain: rice, chicken, and beans are used for training while meat, fish, car, and chair are used for testing (meat and fish as unknown ID; car and chair as unknown OOD).

We define the problem of domain identification as a classification task. Let $x \in X$ be the data point or image from the available dataset and $y \in Y$ be the distinctive label in the dataset. We can define that label $Y$ consists of labels in the domain of interest and labels in the domain outside of the interest that can be denoted as $Y = Y^{ID} \cup Y^{OOD}, Y^{ID} \cap Y^{OOD} = \emptyset$ where ID is the domain of interest and OOD is the domain outside of interest where no similar labels are shared for both datasets.

Since we train the model with ID and  OOD and test with the samples of ID and OOD outside of the training distribution, we need to separate the classes of ID for training and testing such that the classes are different for train and test, similarly for OOD. The ID classes are splittted into $Y^{ID}_{train}$ and $Y^{ID}_{test}$ where $Y^{ID}_{train} \cap Y^{ID}_{test} = \emptyset$. OOD classes can be divided into $Y^{OOD}_{train}$ and $Y^{OOD}_{test}$ where $Y^{OOD}_{train} \cap Y^{OOD}_{test} = \emptyset$

Then, the dataset can be defined as a set of the tuple of data points and classes denoted as $D = \{ (x, y) | x \in X, y \in Y \}$. The dataset in the domain of interest can be denoted as $D^{ID} = \{ (x, y) \in D | y \in Y^{ID} \}$ while the dataset in the domain outside of the interest can be denoted as $D^{OOD} = \{ (x, y) \in D | y \in Y^{OOD} \}$

Each $D^{ID}$ and $D^{OOD}$ have the subset of train and test that is defined in Equation \ref{eq:id_train} and \ref{eq:ood_train} respectively.

\begin{equation}
  \begin{split}
  & D^{ID}_{train} = \{ (x, y) \in D^{ID}  | y \in Y^{ID}_{train} \} \\
  & D^{ID}_{test} = D^{ID} \backslash D^{ID}_{train}
  \end{split}
  \label{eq:id_train}
\end{equation}

\begin{equation}
  \begin{split}
  & D^{OOD}_{train} = \{ (x, y) \in D^{OOD}  | y \in Y^{OOD}_{train} \} \\
  & D^{OOD}_{test} = D^{OOD} \backslash D^{OOD}_{train}
  \end{split}
  \label{eq:ood_train}
\end{equation}

The approximation function for domain identification can be seen in Equation \ref{eq:approx_func}. The $\bar{x}$ is a sample outside of the training distribution, different classes, either ID or OOD. The number 0 represents the OOD while the number 1 represents the ID. 

In the closed-world scenario, during prediction, the samples are taken from the same training distribution. We emphasize the $\bar{x}$ to indicate that the image or data points are different from the training distribution. In other words, the classes are not similar with $Y^{ID}_{train}$ and $Y^{OOD}_{train}$. 

\begin{equation}
    f: \bar{x} \rightarrow \{0, 1\}
\label{eq:approx_func}
\end{equation}

\subsection{Evaluation Protocol}

\begin{figure*}[]
  \centering
  \includegraphics[scale=0.9]{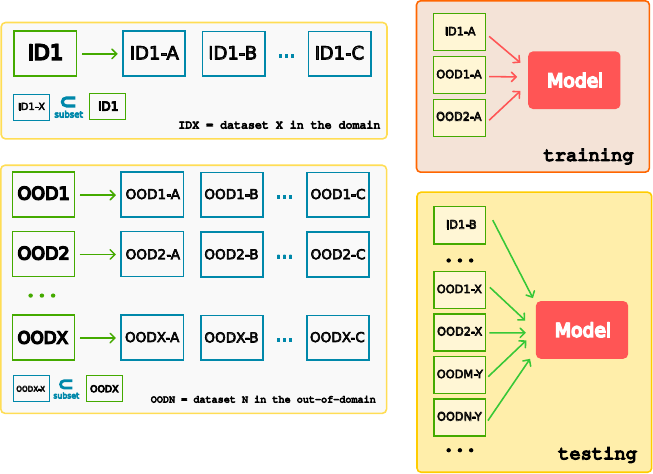}
  \caption{Evaluation model for domain identification for unknown classes in supporting open-world recognition for domain-specific tasks.}
  \label{fig:evaluation}
\end{figure*}

Many evaluation processes have been undertaken by OWR that mostly emphasize its capability to distinguish between known classes and unknown classes and how robust the model to learn continually \cite{bendale2015_owr, fontanel2020_owr_clustering, bukhari2022_owr}. Accuracy \cite{guo_lightweight_2024, fontanel2020_owr_clustering, mancini2019_deep_owr, guo2019_cascades_owr} and F-Measure \cite{xu_2019_owl_product, WANG2023119054, guo2019_cascades_owr} are the common evaluation for open-world recognition. Meanwhile, balanced accuracy has been employed in measuring unknown rejection capability in open-set recognition \cite{schlachter2020_osr_intraclass, Alfarisy2023QuadChannelCP}. 

Throughout our experimentation, we used the Balanced Accuracy (BACCU) \cite{Brodersen2010TheBA} metrics to assess the model in separating unknown ID and OOD which is shown in Equation \ref{eq:baccu}. The metrics take two classes: positive and negative labels. In this problem, positive samples denote the unknown ID while negative samples denote the unknown OOD. $P$ and $N$ are the total numbers of ID and OOD in the test set respectively. $TP$ is the true positive prediction while $TN$ is the true negative prediction. Using this formulation, the proportional score for each label prediction can be obtained.

\begin{equation}
     BACCU = \frac{1}{2} (\frac{TP}{P} + \frac{TN}{N})
    \label{eq:baccu}
\end{equation}

The BACCU yields a score using predictions from different sets of classes from the train set as depicted in Figure \ref{fig:evaluation}.  The dataset in the domain and outside of the domain need to be prepared first and denoted as ID and OOD respectively. From these datasets, the new dataset can be constructed which takes the subset of the original dataset that can be denoted as $A, B, ..., N$. In each dataset, a dataset with the letter $A$, either ID or OOD has a different set of classes than other generated datasets in the same original dataset. For instance, $ID_1 A$ has different labels from $ID_1 B$. During training, The dataset ID (positive labels) and the other two datasets of OOD with different domains are employed as negative labels. In the testing phase, the evaluation has to be in different classes to asses its robustness against the unknown classes in ID or OOD. Hence, a different letter is used for ID. In other words, the same dataset with different classes is being tested. For OOD, we use different datasets as well as the same dataset with different sets of classes. In figure \ref{fig:evaluation}, $X$ denotes the classes that are different from training (different from $A$) and $Y$ can be any set of classes. The $m$ and $n$ denote different domains from the training set of OOD (different from 2 and 3 as shown in the training phase).

\subsection{Methods}
We investigated the feature utilization from ImageNet weights in several deep-learning architectures using three approaches. The first approach is by learning the latent features by linear classifier using fully connected layers. The second approach is through AutoML systems (AutoSkLearn 2.0). Using this approach, we will obtain the optimal ensembles of machine learning models which becomes a good baseline for unknown separation. Another approach is using clustering followed by a Nearest Class Mean (NCM) classifier. FINCH algorithm \cite{Sarfraz_2019_CVPR} was used to obtain the clusters of the training dataset. Using FINCH, the number of clusters is not explicitly specified which could help find the appropriate representable clusters specifically in the feature that has no semantic attributes with the domain of interest. FINCH works by using the the relation of first neighbor which is a collection of samples that becomes a cluster when each data point has an adjacency to the other. The neighbor is calculated using Equation \ref{eq:finch} \cite{Sarfraz_2019_CVPR} and the adjacency link matrix is constructed where $k^1_i$ denotes the first neighbor of data point $i$. After the first clusters are established, FINCH will iteratively merge the clusters by using the distance of the centroid of the clusters found. Hence, FINCH produces one or more partitions.

\begin{equation}
    A(i, j) = 
    \begin{cases}
        1 &  \text{if} \: j = k^1_i \: \text{or} \: k^1_j = 1 \: \text{or} \: k^1_i = k^1_j\\
        0 &  otherwise
    \end{cases}
\label{eq:finch}   
\end{equation}

Since FINCH produces several partitions consisting of several clusters, we choose the largest number of clusters in the partition. We assume that a larger number of clusters will provide a more precise representation of both ID and OOD.  Then the centroids or means from these clusters were obtained and utilized in NCM to classify the unknown classes.

In AutoML systems, we need to specify the budget in seconds or iterations. In our case, we set it to 30 minutes for AutoSklearn 2.0. The AutoML will automatically choose good hyperparameters and machine learning methods. In Nearest Class Mean (NCM) classifier, FINCH was used to cluster the data points by sequentially finding the cluster from $D^{ID}_{train}$ followed by $D^{OOD}_{train}$. Then, the NCM classifier takes the mean as a centroid of the clusters for the classification method. A class (ID and OOD) that is close to the centroid will be assigned to the new data point.

\section{Dataset}
In investigating our problem, we need to determine various domains. We use the available dataset to represent the domain that is not considered too small for representing each class. We select garbage, food, dogs, plants, and birds as the domains that we experiment with. Most datasets, except the garbage domain, have a high number of labels. Hence, we pick 20 classes from the raw dataset to construct two or three different datasets in the same domain to support our experimentation. We listed the classes that were employed in our study as presented in Tabel \ref{tbl:list_labels}

\begin{table}[]
\caption{The list of the labels for each constructed dataset.}
\label{tbl:list_labels}
\centering
\begin{tabular}{lp{12.5cm}}
\toprule
Dataset           & Classes/Labels                                                                                                                                                                                                                                                                                                                                                                                                          \\ 
\midrule
Garbage6          & cardboard, glass, metal, paper, plastic, trash                                                                                                                                                                                                                                                                                                                                                                          \\
Garbage7          & battery, biological, brwon-glass, clothes, green-glass, shoes, trash                                                                                                                                                                                                                                                                                                                                                    \\
Uecfood\_20A      & eels on rice, pilaf, chicken-'n'-egg on rice, pork cutlet on rice, sushi, chicken rice, fried rice, tempura bowl, bibimbap, croissant, roll bread, raisin bread, chip butty, hamburger, pizza, sandwiches, udon noodle, tempura udon, soba noodle, beef noodle                                                                                                                                                          \\
Uecfood\_20B      & tensin noodle, fried noodle, spaghetti, Japanese-style pancake, takoyaki, gratin, sauteed vegetables, croquette, grilled eggplant, sauteed spinach, vegetable tempura, potage, sausage, oden, omelet, ganmodoki, jiaozi, stew, teriyaki grilled fish, fried fish                                                                                                                                                        \\
Uecfood\_20C      & grilled salmon, salmon meuniere , sashimi, grilled pacific saury , sukiyaki, sweet and sour pork, lightly roasted fish, steamed egg hotchpotch, tempura, fried chicken, sirloin cutlet , nanbanzuke, boiled fish, seasoned beef with potatoes, hambarg steak, beef steak, dried fish, ginger pork saute, spicy chili-flavored tofu, yakitori                                                                            \\
StanfordDogs\_20A & black and tan coonhound, redbone, italian greyhound, scottish deerhound, bedlington terrier, lakeland terrier, australian terrier, boston bull, vizsla, english setter, clumber, english springer, sussex spaniel, border collie, bernese mountain dog, eskimo dog, newfoundland, cardigan, miniature poodle, dingo                                                                                                     \\
StanfordDogs\_20B & japanese spaniel, afghan hound, bluetick, borzoi, otterhound, border terrier, cairn, dandie dinmont, scotch terrier, curly coated retriever, chesapeake bay retriever, gordon setter, kuvasz, shetland sheepdog, collie, bouvier des flandres, german shepherd, bull mastiff, tibetan mastiff, samoyed                                                                                                                  \\
StanfordDogs\_20C & blenheim spaniel, papillon, irish wolfhound, kerry blue terrier, norfolk terrier, airedale, silky terrier, lhasa, brittany spaniel, irish water spaniel, groenendael, komondor, entlebucher, boxer, saint bernard, pug, leonberg, keeshond, brabancon griffon, mexican hairless                                                                                                                                         \\
VnPlant\_20A      & excoecaria sp, terminalia catappa, ocimum basilicum, bischofia trifoliata, excoecaria cochinchinensis, ficus auriculata, mallotus barbatus, ageratum conyzoides, capsicum annuum, agave americana, ocimum sanctum, calotropis gigantea, baccaurea sp, costus speciosus, acanthus integrifolius, senna alata, clycyrrhiza uralensis fish, callerya speciosa, piper betle, dimocarpus longan                              \\
VnPlant\_20B      & microcos tomentosa, euphorbia pulcherrima, averrhoa carambola, euphorbia hirta, sanseviera canaliculata carr, gonocaryum lobbianum, crinum asiaticum, quisqualis indica, stachytarpheta jamaicensis, platycladus orientalis, nelumbo nucifera, passiflora foetida, gynura divaricata, cymbopogon, caprifoliaceae, dianella ensifolia, barleria lupulina, eichhoriaceae crassipes, eleutherine bulbosa, ruellia tuberosa \\
VnPlant\_20C      & maesa, rhinacanthus nasutus, artocarpus heterophyllus, vernonia amygdalina, lantana camara, abrus precatorius, hibiscus sabdariffa, spondias dulcis, justicia gendarussa, euphorbia tithymaloides, hibiscus rosa sinensis, cleistocalyx operculatus, amomum longiligulare, rauvolfia tetraphylla, glycosmis pentaphylla, aloe vera, paederia lanuginosa, plukenetia volubilis, solanum torvum, ixora coccinea           \\
Birds\_20A        & abbotts babbler, abbotts booby, abyssinian ground hornbill, african crowned crane, african emerald cuckoo, african firefinch, african oyster catcher, african pied hornbill, albatross, alberts towhee, alexandrine parakeet, alpine chough, altamira yellowthroat, american avocet, american bittern, american coot, american flamingo, american goldfinch, american kestrel, american pipit                           \\
Birds\_20B        & american redstart, american wigeon, amethyst woodstar, andean goose, andean lapwing, andean siskin, anhinga, anianiau, annas hummingbird, antbird, antillean euphonia, apapane, apostlebird, araripe manakin, ashy storm petrel, ashy thrushbird, asian crested ibis, asian dollard bird, auckland shaq, austral canastero                                                                                              \\
Birds\_20C        & australasian figbird, avadavat, azaras spinetail, azure breasted pitta, azure jay, azure tanager, azure tit, baikal teal, bald eagle, bald ibis, bali starling, baltimore oriole, bananaquit, band tailed guan, banded broadbill, banded pita, banded stilt, bar-tailed godwit, barn owl, barn swallow                                                                                                                  \\ 

\bottomrule
\end{tabular}
\end{table}

\subsection{Garbage Domain}

In the garbage domain, we took two datasets from \textit{Kaggle} which consist of 6 and 12 classes. For the 12 classes of the garbage dataset, we found that some classes are similar to the six classes of the dataset. Hence, we employed seven classes only which are battery, biological, brown-glass, clothes, green-glass, shoes, and trash. We denote the six classes with Garbage6 for simplicity that can be downloaded through \cite{garbage6} while we denote the seven classes with Garbage7 that can be downloaded through \cite{garbage12}. 

\subsection{Food Domain}

We use the dataset from \cite{matsuda12} that contains 100 labels of food called UECFOOD100. We split the dataset to construct three datasets by taking 20 classes namely Uecfood\_20A, Uecfood\_20B, and Uecfood\_20C. The detailed information about the labels is presented in Table \ref{tbl:list_labels}.

Before splitting, we investigated the class distribution of each label in this dataset and found a significantly imbalanced set. Hence, we remove the classes that are considered high before proceeding to the data split. The labels that have been removed: 1 (rice), 6 (beef curry), 12 (toast), 23 (ramen noodle), 36 (miso soup), 68 (egg sunny-side up), and 87 (green salad).

\subsection{Dogs Domain}
We took the Stanford Dogs Dataset for the dogs' domain which consists of 120 species of dogs from around the world \cite{KhoslaYaoJayadevaprakashFeiFei_FGVC2011}. We made up three datasets from this by taking 20 classes each which we denote as StanfordDogs\_20A, StanfordDogs\_20B, and StanfordDogs\_20C. More information about the labels is shown in Table \ref{tbl:list_labels}.

\subsection{Plants Domain}

We took the Vietnamese medicinal plant images to represent the plant domains \cite{quoc_vnplant-200_2020}. We use high-resolution images with 512 x 512 pixels. This dataset has 200 species. We construct three datasets by taking 20 classes which we denoted as VnPlants\_20A, VnPlants\_20B, and VnPlants\_20C. The labels of the VnPlants used in this study is listed in Table \ref{tbl:list_labels}.

\begin{table}[ht]
\caption{Experimentation scenario for five different domains. The (+) sign shows the ID dataset while the (-) sign shows the OOD dataset. The postfix "\_20N" shows the subset of the raw dataset that contains 20 classes and "N" shows its distinction to other datasets in the same domain.}
\label{tbl:scenario}
\centering
\footnotesize
\begin{tabular}{p{2cm}p{6.4cm}p{7cm}}
\toprule
\multicolumn{1}{p{2cm}}{Experiment's name (domain)} & Training                                           & Testing                                           \\ 
\midrule
EXP1 (garbage)                 & (+) Garbage6 (-) Uecfood\_20A, StanfordDogs\_20A   & (+) Garbage7 (-) Uecfood\_20B, StanfordDogs\_20B, VnPlant\_20B, Birds\_20B        \\
EXP2 (food)                    & (+) Uecfood\_20A (-) Garbage6, StanfordDogs\_20A   & (+) Uecfood\_20B, Uecfood\_20C (-) Garbage7, StanfordDogs\_20B, VnPlant\_20B, Birds\_20B     \\
EXP3 (dogs)                    & (+) StanfordDogs\_20A (-) Uecfood\_20A, Birds\_20A & (+) StanfordDogs\_20B, StanfordDogs\_20C (-) Uecfood\_20B, Birds\_20B, Garbage7, VnPlant\_20B \\
EXP4 (plants)                  & (+) VnPlant\_20A (-) Uecfood\_20A, Birds\_20A      & (+) VnPlant\_20B, VnPlant\_20C (-) Uecfood\_20B, Birds\_20B, Garbage7, StanfordDogs\_20B \\
EXP5 (birds)                   & (+) Birds\_20A (-) Garbage6, StanfordDogs\_20A     & (+) Birds\_20B, Birds\_20C (-) Garbage7, StanfordDogs\_20B, Uecfood\_20B, VnPlant\_20B     \\ 
\bottomrule
\end{tabular}
\end{table}

\subsection{Birds Domain}


We use the dataset from the \cite{birds400}  to represent the bird domain. This dataset has various species with a large number of classes. We also picked three subsets of 20 classes to construct two datasets namely Birds\_20A, Birds\_20B, and Birds\_20C. The list of the labels in the Birds dataset is presented in Table \ref{tbl:list_labels}.

\section{Experimentation and Discussion}

In this section, we reported the results using the proposed evaluation process with three different approaches as discussed in the method section. We also analyze the effect of different pre-trained models on the performance of unknown class domain identification. We evaluated the approaches using five different domains: garbage, food, dogs, plants, and birds.

The domain identification performance is manifested through designed experiments where each represents a domain of interest that is presented in Table \ref{tbl:scenario}. The first column shows the names of experiments that represent the domain. For instance, in EXP1, the domain of interest is garbage. An agent in this experiment only tackles garbage-related problems. In other words, the agent is interested in identifying unknown classes in the garbage domain only. The second and third columns are the training sets that consist of the reference of unknown ID (+) and unknown OOD (-). We make sure that the labels in training with the testing are different. As in nature, the model will encounter any domain of unknown mimicking a reliable scenario (e.g., In EXP1, food and dog samples, were used as the negative domain for training while plants and bird samples were included during testing that represent unknown OOD).

\subsection{Transfer Learning}

We employ the protocol for transfer learning with various pre-trained models as a feature extractor that is presented in Table \ref{tbl:fc_results}. The "mnet\_v3\_large" is the MobileNet version 3 large model \cite{Howard2019}, "efficientnet\_v2\_l" is the EfficientNet version 2 large model \cite{pmlr_v139_tan21a}, resnet152 is the popular ResNet architecture \cite{He_2016_CVPR}, wide\_resnet101\_2 is the wide residual networks \cite{Zagoruyko2016WideRN}, vit\_b\_16 is the Vision Transformer base model \cite{Dosovitskiy2020AnII}, and swin\_v2\_b is the Swin Transformer version 2 model \cite{Liu_2022_CVPR}.

We evaluated with two different linear classifiers which are FC1 and FC2. FC1 is the classifier that connects directly to the backbone of the pre-trained model (output layer) while FC2 is the classifier that uses intermediate layers between the backbone and output layers which we set to 256 followed by swish activation function \cite{Ramachandran2017SwishAS}. The optimal number of neurons of intermediate layers can be studied in the future depending on the domain's problem.

We can observe the BACCU score from Table \ref{tbl:fc_results} that some models achieved outstanding performance. In \break mnet\_v3\_large, FC2 provided the BACCU score higher than 90\% in the dogs and plants domain. In efficientnet\_v2\_l, poor performance was noticed when attaching the output layers directly to the backbone. However, with FC2, the performance increased significantly which could achieve above 80\% in terms of dogs and plants domain. In resnet152, almost in all domains, either FC1 or FC2 provided a BACCU score above 80\%.  Similar performance is also experienced by wide\_resnet101\_2. However, in plants domain wide\_resnet101\_2 with FC1 yielded higher BACCU score than FC2 indicating that intermediate layers in this case do not help to separate unknown better. In vit\_b\_16, the superior performance was obtained in the food, plants, and birds domain in which both FC1 and FC2 resulted in the BACCU score above roughly 97\%. In swin\_v2\_b, the classifier provided approximately higher than 90\% in terms of food, plants, and birds domain. In the swin transformer case, FC1 provided higher performance than FC2 for the garbage, food, and dog domains.

The BACCU scores also indicate that a pre-trained model has a propensity to excel in identifying a particular domain as shown in the bold style text. Using FC2, mnet\_v3\_large provided the best BACCU score in terms of garbage and dog domain surpassing other complex architecture. Using FC1, vit\_b\_16 yielded the best BACCU score in terms of good and plant domains. While using FC2, vit\_b\_16 can result in the highest BACCU score for the bird domain.

\begin{table}[]
\centering
\caption{BACCU score using a transfer learning approach with a different feature through the pre-trained model. The backbone parameters were frozen and behaved as the feature extractor.}
\label{tbl:fc_results}
\begin{tabular}{clrr}
\toprule
\multicolumn{1}{l}{Pre-trained Model} & Domain         & \multicolumn{1}{l}{FC1} & \multicolumn{1}{l}{FC2} \\ 
\midrule
 \multirow{5}{*}{\textbf{mnet\_v3\_large}}                      & \textbf{EXP1 (garbage)} & 0.8407                  & \textbf{0.8516}                  \\
                                     & EXP2 (food)    & 0.8355                  & 0.8577                  \\
                                     & \textbf{EXP3 (dogs)}    & 0.9212                  & \textbf{0.9429}                  \\
                                     & EXP4 (plants)  & 0.8886                  & 0.9088                  \\
                                     & EXP5 (birds)   & 0.8134                  & 0.8711                  \\ \hline
\multirow{5}{*}{efficientnet\_v2\_l}                  & EXP1 (garbage) & 0.5759                  & 0.6699                  \\
                                     & EXP2 (food)    & 0.6131                  & 0.7989                  \\
                                     & EXP3 (dogs)    & 0.6071                  & 0.8279                  \\
                                     & EXP4 (plants)  & 0.6241                  & 0.8150                  \\
                                     & EXP5 (birds)   & 0.6150                  & 0.7513                  \\ \hline
\multirow{5}{*}{resnet152}                            & EXP1 (garbage) & 0.7245                  & 0.7321                  \\
                                     & EXP2 (food)    & 0.8042                  & 0.8152                  \\
                                     & EXP3 (dogs)    & 0.8713                  & 0.8879                  \\
                                     & EXP4 (plants)  & 0.8952                  & 0.8971                  \\
                                     & EXP5 (birds)   & 0.8897                  & 0.8933                  \\ \hline
\multirow{5}{*}{wide\_resnet101\_2}                   & EXP1 (garbage) & 0.7213                  & 0.7163                  \\
                                     & EXP2 (food)    & 0.8073                  & 0.8258                  \\
                                     & EXP3 (dogs)    & 0.8690                  & 0.8911                  \\
                                     & EXP4 (plants)  & 0.9365                  & 0.9148                  \\
                                     & EXP5 (birds)   & 0.8644                  & 0.8833                  \\ \hline
\multirow{5}{*}{\textbf{vit\_b\_16}}                           & EXP1 (garbage) & 0.7458                  & 0.7671                  \\
                                     & \textbf{EXP2 (food)}    & \textbf{0.9793}                  & 0.9784                  \\
                                     & EXP3 (dogs)    & 0.8703                  & 0.8793                  \\
                                     & \textbf{EXP4 (plants)}  & \textbf{0.9967}                  & 0.9963                  \\
                                     & \textbf{EXP5 (birds)}   & 0.9746                  & \textbf{0.9790}                  \\ \hline
\multirow{5}{*}{swin\_v2\_b}                          & EXP1 (garbage) & 0.7186                  & 0.6501                  \\
                                     & EXP2 (food)    & 0.9645                  & 0.9522                  \\
                                     & EXP3 (dogs)    & 0.7811                  & 0.7809                  \\
                                     & EXP4 (plants)  & 0.9851                  & 0.9874                  \\
                                     & EXP5 (birds)   & 0.9293                  & 0.9321                  \\ 
\bottomrule
\end{tabular}
\end{table}

\subsection{Automated Machine Learning (AutoML)}

As an alternative to the standard transfer learning approach, we experimented with the AutoML model that will result in optimal ensembled machine learning models to identify unknown classes in the domain of interest. This may avoid a biased justification in determining the classifier part for different pre-trained models. The BACCU scores are presented in Table \ref{tbl:automl_results}.


\begin{table}[]
\centering
\caption{BACCU score using AutoML model with different features through the pre-trained model.}
\label{tbl:automl_results}
\begin{tabular}{clr}
\toprule
\multicolumn{1}{l}{Pre-trained Model}     & Domain                  & \multicolumn{1}{l}{BACCU} \\
\midrule
\multirow{5}{*}{\textbf{mnet\_v3\_large}} & \textbf{EXP1 (garbage)} & \textbf{0.8595}           \\
                                          & EXP2 (food)             & 0.8825                    \\
                                          & \textbf{EXP3 (dogs)}    & \textbf{0.9583}           \\
                                          & EXP4 (plants)           & 0.9482                    \\
                                          & EXP5 (birds)            & 0.8826                    \\ \hline
\multirow{5}{*}{efficientnet\_v2\_l}      & EXP1 (garbage)          & 0.6835                    \\
                                          & EXP2 (food)             & 0.8379                    \\
                                          & EXP3 (dogs)             & 0.8607                    \\
                                          & EXP4 (plants)           & 0.8719                    \\
                                          & EXP5 (birds)            & 0.8156                    \\ \hline
\multirow{5}{*}{resnet152}                & EXP1 (garbage)          & 0.7183                    \\
                                          & EXP2 (food)             & 0.8772                    \\
                                          & EXP3 (dogs)             & 0.9117                    \\
                                          & EXP4 (plants)           & 0.9470                    \\
                                          & EXP5 (birds)            & 0.9409                    \\ \hline
\multirow{5}{*}{wide\_resnet101\_2}       & EXP1 (garbage)          & 0.7354                    \\
                                          & EXP2 (food)             & 0.8743                    \\
                                          & EXP3 (dogs)             & 0.9004                    \\
                                          & EXP4 (plants)           & 0.9614                    \\
                                          & EXP5 (birds)            & 0.9145                    \\ \hline
\multirow{5}{*}{\textbf{vit\_b\_16}}      & EXP1 (garbage)          & 0.7388                    \\
                                          & EXP2 (food)             & 0.9640                    \\
                                          & EXP3 (dogs)             & 0.8670                    \\
                                          & EXP4 (plants)           & 0.9883                    \\
                                          & \textbf{EXP5 (birds)}   & \textbf{0.9791}           \\ \hline
\multirow{5}{*}{\textbf{swin\_v2\_b}}     & EXP1 (garbage)          & 0.7109                    \\
                                          & \textbf{EXP2 (food)}    & \textbf{0.9694}           \\
                                          & EXP3 (dogs)             & 0.7966                    \\
                                          & \textbf{EXP4 (plants)}  & \textbf{0.9952}           \\
                                          & EXP5 (birds)            & 0.9410                    \\
                                          
\bottomrule
\end{tabular}
\end{table}

From Table \ref{tbl:automl_results}, we can notice that most of the models can provide the BACCU score above 90\% for a particular domain. Using mnet\_v3\_large, the BACCU scores of approximately 96\% and 95\% were obtained for the dogs and plant domains respectively. Unfortunately, using efficientnet\_v2\_l, the BACCU scores were not reaching 90\% in all domains. The best score was achieved for the plant's domain at about 87\% of the BACCU score. For resnet152, the AutoML model provided scores roughly 91\%, 95\%, and 94\% for dogs, birds, and plants domains respectively. In wide\_resnet101\_2, the model achieved roughly 92\% and 96\% of BACCU score for birds and plants domain respectively. In vit\_b\_16, AutoML yielded approximately 96\%, 98\%, and 99\% of BACCU score in terms of food, birds, and plants domain respectively. The performance through vision transformer features is almost flawless in these domains. For swin\_v2\_b, roughly 94\%, 97\%, and almost 100\% were achieved by the model for birds, food, and plants domains.

We also observe that particular pre-trained models provided an excellent performance for a particular domain. For the garbage and dog domain, mnet\_v3\_large yielded the best performance compared to other pre-trained models. For the birds domain, vit\_b\_16 produced the highest BACCU score. Meanwhile, in the food and plants domain, AutoML with swin\_v2\_b provided the best performance.

\subsection{NCM Classifier with FINCH}

In addition to previous approaches, we investigated the utilization of natural features by taking the mean of each ID and OOD dataset (NCM) and by calculating the centroids from clusters generated from FINCH (NCM+FINCH). This requires a simple decision based on the nearest distance to a centroid/mean. Unlike the two previous methods that need a training mechanism that transforms the natural features into different vector spaces. Using this approach, we can observe the effect of raw features by a simple classifier which could unveil the capability of the raw representation in the pre-trained models in terms of identification of the domain of unknown classes.

The BACCU scores are presented in Table \ref{tbl:ncm_results}. Overall, NCM with FINCH produced a higher score than NCM alone in most cases. In mnet\_v3\_large, FINCH improved the NCM in terms of garbage, food, and bird domain. In efficientnet\_v2\_l, using FINCH improved the NCM prediction by a large margin. For instance, in the plant domain, FINCH improved the NCM by roughly 13\% of the BACCU score. For resnet152 and wide\_resnet101\_2, almost all domain problems were improved using FINCH, except for the dog domain. In swin\_v2\_b, most of the domain identifications were improved by FINCH. Meanwhile, using vit\_b\_16, the results were competitive (except in the dogs' domain in which FINCH enhanced the NCM by approximately 12\%).

In many circumstances, NCM produced a higher score than NCM with FINCH. For convolutional-based networks, using NCM only provided greater performance than NCM with FINCH. Other than that circumstance, the results where NCM yielded higher scores than NCM with FINCH were considered competitive as shown in vit\_b\_16 and swin\_v2\_b.

We can also observe that mnet\_v3\_large and vit\_b\_16 can produce the highest BACCU score in identifying unknown ID. In terms of garbage domain, mnet\_v3\_large yielded the highest BACCU score roughly 78\%. Meanwhile, for other domains, vit\_b\_16 produced almost flawless capability as shown in a great BACCU score which is higher than 92\%. For example, in the plant domain, almost 100\% of the BACCU score was achieved.

\begin{table}[]
\centering
\caption{BACCU score using NCM classifier based on the clusters found from FINCH with different features through the pre-trained model.}
\label{tbl:ncm_results}
\begin{tabular}{clrr}
\toprule
\multicolumn{1}{l}{Pretrained Model}      & Domain                  & \multicolumn{1}{l}{NCM} & \multicolumn{1}{p{1cm}}{NCM +FINCH} \\ 
\midrule
\multirow{5}{*}{\textbf{mnet\_v3\_large}} & \textbf{EXP1 (garbage)} & 0.7748                  & \textbf{0.7780}               \\
                                          & EXP2 (food)             & 0.7096                  & 0.7790                        \\
                                          & EXP3 (dogs)             & 0.8127                  & 0.7996                        \\
                                          & EXP4 (plants)           & 0.8131                  & 0.8119                        \\
                                          & EXP5 (birds)            & 0.6963                  & 0.7688                        \\ \hline
\multirow{5}{*}{efficientnet\_v2\_l}      & EXP1 (garbage)          & 0.5503                  & 0.5953                        \\
                                          & EXP2 (food)             & 0.5446                  & 0.6495                        \\
                                          & EXP3 (dogs)             & 0.5285                  & 0.5853                        \\
                                          & EXP4 (plants)           & 0.5710                  & 0.7056                        \\
                                          & EXP5 (birds)            & 0.5733                  & 0.6616                        \\ \hline
\multirow{5}{*}{resnet152}                & EXP1 (garbage)          & 0.6852                  & 0.7143                        \\
                                          & EXP2 (food)             & 0.7058                  & 0.7697                        \\
                                          & EXP3 (dogs)             & 0.8125                  & 0.7427                        \\
                                          & EXP4 (plants)           & 0.7557                  & 0.8569                        \\
                                          & EXP5 (birds)            & 0.7488                  & 0.8687                        \\ \hline
\multirow{5}{*}{wide\_resnet101\_2}       & EXP1 (garbage)          & 0.6799                  & 0.7168                        \\
                                          & EXP2 (food)             & 0.7024                  & 0.7611                        \\
                                          & EXP3 (dogs)             & 0.8093                  & 0.7283                        \\
                                          & EXP4 (plants)           & 0.8144                  & 0.8438                        \\
                                          & EXP5 (birds)            & 0.7758                  & 0.8496                        \\ \hline
\multirow{5}{*}{\textbf{vit\_b\_16}}      & EXP1 (garbage)          & 0.7090                  & 0.7057                        \\
                                          & \textbf{EXP2 (food)}    & \textbf{0.9810}         & 0.9808                        \\
                                          & \textbf{EXP3 (dogs)}    & 0.8157                  & \textbf{0.9399}               \\
                                          & \textbf{EXP4 (plants)}  & \textbf{0.9971}         & 0.9905                        \\
                                          & \textbf{EXP5 (birds)}   & 0.9680                  & \textbf{0.9743}               \\ \hline
\multirow{5}{*}{swin\_v2\_b}              & EXP1 (garbage)          & 0.6676                  & 0.7000                        \\
                                          & EXP2 (food)             & 0.9444                  & 0.9726                        \\
                                          & EXP3 (dogs)             & 0.7056                  & 0.8341                        \\
                                          & EXP4 (plants)           & 0.9928                  & 0.9897                        \\
                                          & EXP5 (birds)            & 0.8314                  & 0.9592                        \\ 
                                          
\bottomrule
\end{tabular}
\end{table}

\subsection{Analysis Across the Approaches}

We performed analysis through the empirical evidence that each approach produced. We analyzed the BACCU score across approaches by observing the highest performance and the effect of pre-trained models on a particular domain.

The highest score for garbage was obtained by using the AutoML model with a feature through mnet\_v3\_large. Interestingly, the BACCU score was about 86\% surpassing other pre-trained models even compared to vit\_b\_16 which has a larger number of parameters than mnet\_v3\_large. We noticed that the AutoML model was competitive with the transfer learning approach by only 0.8\% higher. Meanwhile compared to NCM with FINCH, the AutoML yielded significant enhancement by approximately 9\%. We also observe that mnet\_v3\_large consistently produced the highest BACCU score across the approaches suggesting that mnet\_v3\_large suitable for identifying garbage domain.

The highest score for the food domain was achieved by NCM using vit\_b\_16 features. The BACCU score was obtained at about 97\% through NCM. The vit\_b\_16 provided a good representation for all approaches. In NCM and transfer learning approaches, vit\_b\_16 provided the highest BACCU score. In the AutoML approach, vit\_b\_16 was considered competitive with the highest BACCU score from swin\_v2\_b by merely 0.5\% less. A simple approach by using the centroid for each class of dataset (ID and OOD) for unknown domain prediction surprisingly provided a higher score than other approaches. This indicates that a strong representation in the backbone model is more beneficial than the complexity of the classifier in separating unknown ID and OOD.

For the dogs domain, mnet\_v3\_large with AutoML model produced the highest BACCU score approximately 96\%. AutoML provided roughly 2\% higher than the score produced by transfer learning and NCM with FINCH. In the aspect of the pre-trained model, mnet\_v3\_large produced the highest performance for transfer learning and AutoML approaches. Meanwhile, for NCM with FINCH, vit\_b\_16 yielded the highest performance. 

For the plants domain, NCM using the feature from vit\_b\_16 provided the highest performance nearly 100\% of the BACCU score. However, the score was competitive with AutoML and the transfer learning approach which also resulted in nearly 100\% of BACCU score. The vit\_b\_16 consistently provided a strong feature for all approaches and was competitive with swin\_v2\_b in terms of the AutoML model.

For the birds domain, AutoML with the feature through vit\_b\_16 yielded the highest BACCU score across the approaches roughly 98\%. This score was competitive with other approaches and vit\_b\_16 was also observed as the best pre-trained model across the approaches.

\section{Conclusion}
We proposed an evaluation protocol for the domain of interest identification to support open-world recognition in handling domain-specific tasks by incorporating samples in different domains. We investigated the results of the proposed evaluation protocol using three different approaches: linear classifier through transfer learning, AutoML model, and NCM with FINCH. We also investigated the effect of the pre-trained model across approaches for each domain. The empirical results indicate a tendency for a particular pre-trained model to excel in one or more domains of interest. We observed that features derived from MobileNetV3 and ViT-base pre-trained models produced the highest BACCU score for garbage and bird domains respectively across the approaches. Furthermore, a great performance achieved by a simple technique (NCM) suggests that strong representation is imperative for unknown class identification in the domain of interest.











\bibliographystyle{unsrt}  

\bibliography{preprint}



\end{document}